\documentclass[11pt,a4paper]{article}
\usepackage[hyperref]{ranlp2025}
\usepackage[english,russian]{babel}           
\usepackage[T1,T2A,X2]{fontenc}               
\usepackage[utf8]{inputenc}                   

\usepackage{newunicodechar}

\newcommand{\textXtwo}[1]{{\fontencoding{X2}\selectfont #1}}

\newunicodechar{ѣ}{\textXtwo{\cyryat}}
\newunicodechar{Ѣ}{\textXtwo{\CYRYAT}}
\newunicodechar{ѵ}{\textXtwo{\cyrizh}}
\newunicodechar{Ѵ}{\textXtwo{\CYRIZH}}

\usepackage{graphicx}

\newunicodechar{ѧ}{%
  \raisebox{-0.2ex}{\includegraphics[height=1.1ex]{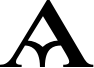}}%
}

\newunicodechar{ѥ}{%
  \raisebox{-0.2ex}{\includegraphics[height=1.1ex]{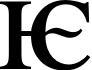}}%
}

\newunicodechar{ѭ}{%
  \raisebox{-0.2ex}{\includegraphics[height=1.1ex]{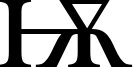}}%
}

\newunicodechar{ѡ}{%
  \raisebox{-0.2ex}{\includegraphics[height=1.1ex]{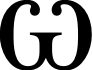}}%
}

\newunicodechar{ѿ}{%
  \raisebox{-0.2ex}{\includegraphics[height=1.1ex]{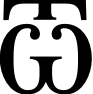}}%
}

\newunicodechar{ꙋ}{%
  \raisebox{-0.2ex}{\includegraphics[height=1.1ex]{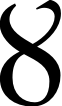}}%
}

\newunicodechar{ꙗ}{%
  \raisebox{-0.2ex}{\includegraphics[height=1.1ex]{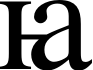}}%
}

\newunicodechar{ѳ}{%
  \raisebox{-0.2ex}{\includegraphics[height=1.1ex]{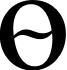}}%
}

\newunicodechar{ѱ}{%
  \raisebox{-0.2ex}{\includegraphics[height=1.1ex]{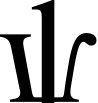}}%
}

\newunicodechar{ѯ}{%
  \raisebox{-0.2ex}{\includegraphics[height=1.1ex]{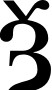}}%
}

\usepackage{microtype}
\usepackage{inconsolata}

\aclfinalcopy 

\title{Evaluating LLMs for Historical Document OCR: A Methodological Framework for Digital Humanities}

\author{Maria Levchenko \\
  Italian Institute of Germanic Studies (IISG), Rome, Italy \\
  University of Bologna, Bologna, Italy \\
  \texttt{marylevchenko@gmail.com} }

\date{}

\begin{document}
\maketitle
\begin{abstract}
Digital humanities scholars increasingly use Large Language Models for historical document digitization, yet lack appropriate evaluation frameworks for LLM-based OCR. Traditional metrics fail to capture temporal biases and period-specific errors crucial for historical corpus creation. We present an evaluation methodology for LLM-based historical OCR, addressing contamination risks and systematic biases in diplomatic transcription. Using 18th-century Russian Civil font texts, we introduce novel metrics including Historical Character Preservation Rate (HCPR) and Archaic Insertion Rate (AIR), alongside protocols for contamination control and stability testing. We evaluate 12 multimodal LLMs, finding that Gemini and Qwen models outperform traditional OCR while exhibiting "over-historicization"—inserting archaic characters from incorrect historical periods. Post-OCR correction degrades rather than improves performance. Our methodology provides digital humanities practitioners with guidelines for model selection and quality assessment in historical corpus digitization.
\end{abstract}

\section{Introduction}
The evolution of large language models (LLMs) into powerful optical character recognition (OCR) tools has created new opportunities in digital humanities, especially for processing historical documents where traditional OCR systems struggle with non-standard typography, evolving orthographic conventions. However, evaluating LLM-based OCR requires fundamentally different methodological approaches than those developed for traditional machine learning systems. Unlike conventional OCR models, where researchers can control training data, modify architectures, and perform fine-tuning, LLMs present unique evaluation challenges: we cannot access their training data, modify their parameters, or retrain them for specific historical corpora. This constraint necessitates new evaluation frameworks that assess and optimize LLM performance through external factors such as prompt engineering, processing modes, and systematic bias detection.

Current OCR evaluation practices prove inadequate for LLM-based historical document processing. Standard metrics like Character Error Rate (CER) and Word Error Rate (WER) fail to capture LLM-specific behaviors such as temporal conflation, where models incorrectly apply orthographic features from different historical periods, or systematic insertion of anachronistic elements. Moreover, the risk of training data contamination, where evaluation texts may have been included in LLM pretraining corpora, undermines traditional benchmarking approaches that assume clean train-test separation.

We address these methodological gaps through a comprehensive evaluation framework, demonstrated via the challenging case study of 18th-century Russian texts printed in Civil font. This domain exemplifies evaluation challenges facing digital humanities: texts feature distinctive orthographic elements (і, ѣ, ъ at word endings), archaic grammatical forms, and syntactic structures unfamiliar to modern readers, while being underrepresented in digital corpora and thus effectively low-resource for LLMs. These linguistic elements are seldom preserved online; even the Russian National Corpus often presents 18th-century texts in post-1918 orthography \cite{savchuk2009corpus}. The diplomatic transcription requirement—preserving exact textual features including line breaks, hyphens, and original typographical errors—further demands precise character-level fidelity that tests LLM capabilities beyond normalized text processing.

Building on recent evidence that LLMs can outperform specialized OCR systems through holistic page processing and prompt engineering \cite{humphries2024, kim2025, sohail2024}, our framework introduces key innovations: (1) contamination-aware dataset creation protocols ensuring evaluation integrity, (2) novel metrics designed to capture LLM behaviors in historical contexts, including Historical Character Preservation Rate (HCPR) and Archaic Insertion Rate (AIR), (3) systematic analysis of processing modes and prompt engineering strategies, (4) comprehensive stability testing accounting for LLM output variability, and (5) feature sensitivity analysis identifying document characteristics that affect performance.

Using this framework, we evaluate 12 leading commercial and open-source multimodal LLMs on a novel dataset of 1,029 pages from 428 unique 18th-century Russian books, revealing systematic patterns in LLM behavior previously undocumented in historical OCR literature. Our analysis uncovers "over-historicization"—a phenomenon where LLMs systematically insert archaic characters eliminated from the target historical period—demonstrating how LLMs exhibit unexpected temporal biases that standard evaluation approaches cannot detect.

\section{Literature review}

Early research found LLM-based OCR often outperforms state-of-the-art pipelines. Multimodal LLMs often transcribe unseen manuscripts zero-shot for printed and even handwritten documents in English, Finnish, Italian and Japanese. For instance, Humphries et al. (2024) report GPT-4–class models achieved CER around 5–7\% on 18th–19th century English manuscripts—a 14\% relative improvement over Transkribus—and further reduced CER to ~1.8\% with LLM-based post-correction \cite{humphries2024}. Similarly, Kim et al. (2025) found general-purpose LLMs outperforming tools like Tesseract and TrOCR on historical tables, and early benchmarks highlight the importance of prompt design (e.g., two-shot prompting, line-by-line input) \cite{kim2025}.

However, recent studies underscore limitations. Crosilla et al. (2025) benchmarked LLMs against Transkribus on multilingual historical datasets and found no consistent overall winner \cite{crosilla2025}. Proprietary models excelled in English, while open-source LLMs and non-English scripts showed weaker performance, reflecting pretraining data biases. Unpredictable generative outputs and hallucinations remain challenges \cite{thomas-2024, boros-2024}. While instruction-tuning can aid post-OCR correction, zero-shot self-correction abilities are still limited.

In summary, while LLMs have advanced OCR for some languages, new risks arise: contamination from training data, unpredictable outputs, and the need for task-specific prompt engineering. Notably, little work has evaluated LLMs on Russian historical texts, motivating our focus.

\section{Methodology and Data Integrity Controls}

\textbf{Preventing Training Data Contamination.} Evaluation of LLMs on OCR tasks is complicated by the risk of test set contamination, as standard benchmarks are often present in LLM pretraining corpora. Prior studies have used n-gram overlap, membership inference attacks (MIAs), and surprisal-based probes, but these methods are limited, especially for historical material \cite{chang-etal-2023-speak, ravichander-etal-2025-information}. To ensure robust evaluation, we created a novel dataset of 18th-century Russian texts, digitized from sources never previously recognized or published, and kept strictly offline during all known LLM pretraining periods.

\textbf{Dataset.} Our corpus consists of 1,029 scanned pages from 428 unique 18th-century books printed in Russian Civil font, sourced from the National Library of Russia's limited-access collection "Русская книга гражданской печати XVIII в. в библиотеках РФ" (Russian Civil Print Books of the 18th Century in Russian Federation Libraries). We stratified the data by publication period (1750--1800), text density, decorative elements, and subject (fiction, science, religion, etc.) to ensure diversity (see Appendix~\ref{sec:dataset} for details). Images below 150ppi were excluded, following evidence of poor LLM OCR performance at low resolution \cite{inoue2025}. 

The ground truth (GT) for this corpus was prepared through a multi-stage process:

\textbf{Layout Analysis}: a YOLOv8 model \cite{10533619}, fine-tuned on a 495-page subset of this corpus, performed region detection; line detection within regions utilized a pre-trained \href{https://huggingface.co/Riksarkivet/rtmdet_lines}{Riksarkivet} model.

\textbf{Initial OCR}: A TrOCR model, also fine-tuned on the same 495-page subset (13,456 lines), generated initial transcriptions for the entire corpus. On a held-out portion of the tool-training data, the TrOCR model achieved a CER of 1.83\%, WER of 7.82\%, and line Exact Match Rate (EMR) of 99.84\%.

\textbf{Manual Correction}: 100\% manual review using the eScriptorium interface. Our transcription adheres to diplomatic principles, preserving period-specific orthography, hyphenation, original errors, typos, and spacing conventions to accurately reflect the source documents. The resulting GT for the entire 1,029-page corpus (which serves as the evaluation set for the LLMs) comprises 28,657 lines and 146,690 words. It will be released upon publication.

GT was produced by a single expert annotator using a two-pass protocol. We audited a stratified sample of 500 lines with a second verifier under the same guidelines; line-level exact-match was 98.6\%, and character-level accuracy was 99.93\%.

\textbf{Baseline: Traditional OCR for Historical Texts.} Traditional OCR systems struggle with 18th-century Russian texts in Civil font due to a confluence of challenges, including visually confusable character pairs (e.g., i/ï, т/ш), divergent historical orthographic conventions, typographic inconsistencies from printer-specific variations, and complex page layouts with decorative elements.

To quantify these difficulties, we tested both a general-purpose OCR system (Tesseract), the multilingual BT5 model from the Surya OCR framework \cite{paruchuri2025surya}, and a specialized “Russian print XVIII cent PyLaia” model trained on similar material via Transkribus (reporting 2.40\% CER on its own data). On a 100-page sample from our dataset, as shown in Table~\ref{tab:ocr}, Surya struggled with the Old Russian orthography (45.96\% CER), PyLaia showed a markedly poorer CER of 26.93\%, and Tesseract performed significantly worse. For reference, a TrOCR model fine-tuned on our data achieved 1.83\% CER—an upper bound, not typical of generic OCR models.

\begin{table}
  \centering
  \begin{tabular}{lccc}
    \hline
    \textbf{OCR System} & \textbf{CER (\%)} & \textbf{WER (\%)} \\
    \hline
    Surya (BT5) & 45.96 & 78.33 \\
    Tesseract OCR 4.0 & 21.55 & 126.10 \\
    Transkribus PyLaia & 26.93 & 29.07 \\
    Fine-tuned TrOCR$^{*}$ & 1.83 & 7.82 \\
    \hline
  \end{tabular}
  \caption{OCR results for Old Russian orthography}
  $^{*}$ Fine-tuned on our dataset; represents an upper bound, not indicative of typical generalization.
  \label{tab:ocr}
\end{table}

This significant performance drop, even for specialized models not fine-tuned on our specific corpus, underscores the generalization limits of traditional OCR and the impracticality of achieving usable results without extensive, resource-intensive retraining for specific collections. Such limitations motivate our investigation into Large Language Models (LLMs) as a more adaptable alternative.

Our study addresses three research questions: \textbf{RQ1:} How do input parameters (processing mode, text density, decorative elements, subject) affect LLM OCR performance for 18th-century Russian Civil font? \textbf{RQ2:} What is the impact of prompt engineering on period-specific orthographic fidelity? \textbf{RQ3:} What are the characteristic error patterns of LLM-based OCR on historical Russian?

\section{Experiment Setup}

We evaluated 12 leading LLMs (see Appendix~\ref{sec:models}), including commercial models (Claude, GPT, Gemini) and open-source models (Llama, Qwen). Models were accessed either via their official APIs or, for open-source models, through the TogetherAI service.

\textbf{Model Stability Evaluation Protocol}. To assess performance consistency, we re-evaluated a subset of our models on a fixed sample of 20 documents daily for seven consecutive days. Stability was measured by the Coefficient of Variation (CV) of daily Word Accuracy scores.

\textbf{Recognition Modes}. \textit{Single Line Processing}: Each text line is processed independently, mirroring traditional OCR. This mode provides minimal context and is efficient but may miss cross-line dependencies.

\textit{Full Page Processing}: The entire page image is provided as a single input, maximizing contextual information. While this may resolve ambiguities, it risks hallucinations or detail loss on dense or complex layouts.

\textit{Sliding Window Processing}: Fixed-size windows (e.g., 3 lines at a time, transcribing the center) provide more context than single-line but may be more robust to local errors than full-page mode.

\textbf{Prompt Engineering Experiments}. We conducted systematic prompt variation experiments (see Appendix~\ref{sec:prompts}), including 1) a baseline prompt with basic image information (“Extract the OCR text from this 18th-century Russian book line. Preserve the original Old Russian orthography.”), 2) context-enhanced prompts in English (including book information and character list), 3) context-enhanced prompts in Russian.

\textbf{Evaluation Metrics}. We employed an evaluation framework with multiple metrics to assess OCR accuracy, historical fidelity, and case sensitivity:

\textit{Standard OCR Metrics}. Character Error Rate (CER) and Word Error Rate (WER), using Levenshtein distance between prediction and ground truth.

\textit{Case-Insensitive Metrics}. CER and WER after lowercasing, to isolate character recognition from case errors.

\textit{Historical Fidelity Metrics}.  Historical Character Preservation Rate (HCPR) for period-specific characters (і/ї, ѣ, ъ); Archaic Insertion Rate (AIR) for insertion of obsolete, pre-Petrine characters.

\textit{Case Preservation Accuracy}. Case Error Rate (CaseER) to specifically assess case assignment errors, with particular focus on visually distinctive characters such as ѣ.

\section{Experiments and Results}
We evaluated all models across the three recognition modes using standard metrics (CER, WER, CI-CER, CI-WER, historical character fidelity, and case accuracy). Table~\ref{tab:model-performance} summarizes model performance for each mode; lower values indicate better performance.

\begin{table*}[ht]
\centering
\begin{tabular}{lccccccc}
\hline
\textbf{Model} & \multicolumn{2}{c}{\textbf{Full page}} & \multicolumn{2}{c}{\textbf{Single Line}} & \multicolumn{2}{c}{\textbf{Sliding Window}} & \\
               & CER (\%) & WER (\%) & CER (\%) & WER (\%) & CER (\%) & WER (\%) & \\
\hline
Gemini-2.5-Pro        & \textbf{3.36} & \textbf{4.69} & 9.35 & 15.99 & 7.83 & 11.77 \\
Gemini-2.5-Flash      & 4.94 & 6.70 & 18.79 & 26.21 & 25.63 & 30.77  \\
Qwen-2.5-VL           & 5.81 & 7.48 & 7.70 & 11.29 & 8.87 & 12.72 \\
Gemini-2.0            & 6.14 & 10.33 & 10.04 & 16.43 & 14.90 & 19.50 \\
Claude-3.5            & 6.79 & 8.46 & \textbf{5.73} & 9.61 & \textbf{7.17} & 11.07 \\
OpenAI-o4-mini        & 6.87 & 9.07 & 9.35 & 13.89 & 8.17 & 11.67 \\
Claude-3.7            & 7.32 & 9.47 & \textbf{5.63} & \textbf{9.13} & 7.35 & \textbf{10.03} \\
GPT-4.1               & 7.90 & 9.76 & 7.55 & 11.89 & 9.59 & 13.35 \\
Llama-4-Maverick      & 8.29 & 11.87 & 8.98 & 16.62 & 11.57 & 16.81 \\
GPT-4o                & 9.23 & 13.66 & 23.75 & 28.30 & 11.93 & 17.13 \\
Llama-4-Scout         & 15.94 & 20.51 & 8.98 & 15.41 & 14.95 & 20.78 \\
\hline
\end{tabular}
\caption{Model Performance Comparison: Character Error Rate (CER) and Word Error Rate (WER) across three recognition modes for each model. Best scores in each column are bolded.}
\label{tab:model-performance}
\end{table*}

\textbf{Recognition Mode Effectiveness}. Table~\ref{tab:fullpage-results} reports performance in full-page mode (the best mode for most models). For each metric, we provide mean values and the observed range (in parentheses) across all documents. Gemini-2.5-Pro achieved the lowest error rates overall. All models showed higher error rates for historical character preservation than for general character recognition, indicating persistent difficulty with period-specific features.
\begin{table*}[ht]
\centering
\begin{tabular}{lcccc}
\hline
\textbf{Model} & \textbf{CER (\%)} & \textbf{WER (\%)} & \textbf{CI-CER (\%)} & \textbf{Hist. Char. Error (\%)} \\
\hline
Gemini-2.5-Pro      & 3.36 (0.14--20.95)   & 4.69 (0.08--31.43)   & 3.19  & 9.83   \\
Gemini-2.5-Flash    & 4.94 (0.75--22.11)   & 6.70 (0.41--22.82)   & 4.81  & 12.86  \\
Qwen-2.5-VL         & 5.81 (0.81--86.86)   & 7.48 (0.99--90.14)   & 5.54  & 16.40  \\
Gemini-2.0          & 6.14 (1.51--22.16)   & 10.33 (1.58--30.55)  & 5.66  & 32.00  \\
Claude-3.5          & 6.79 (0.70--53.96)   & 8.46 (0.00--51.09)   & 5.75  & 15.24  \\
OpenAI-o4-mini      & 6.87 (2.18--57.54)   & 9.07 (2.37--58.85)   & 6.76  & 18.38  \\
Claude-3.7          & 7.32 (0.61--51.40)   & 9.47 (0.21--53.93)   & 6.21  & 15.29  \\
GPT-4.1             & 7.90 (1.20--31.14)   & 9.76 (1.96--31.85)   & 7.80  & 16.94  \\
Llama-4-Maverick    & 8.29 (1.34--72.57)   & 11.87 (1.68--69.81)  & 7.77  & 22.33  \\
GPT-4o              & 9.23 (1.89--40.08)   & 13.66 (1.30--48.87)  & 9.07  & 20.70  \\
Llama-4-Scout       & 15.94 (2.09--97.00)  & 20.51 (1.70--99.18)  & 14.95 & 42.23  \\
\hline
\end{tabular}
\caption{Full page mode results. For CER and WER, ranges in parentheses show minimum and maximum values across all documents.}
\label{tab:fullpage-results}
\end{table*}

\begin{figure}[t]
  \includegraphics[width=\columnwidth]{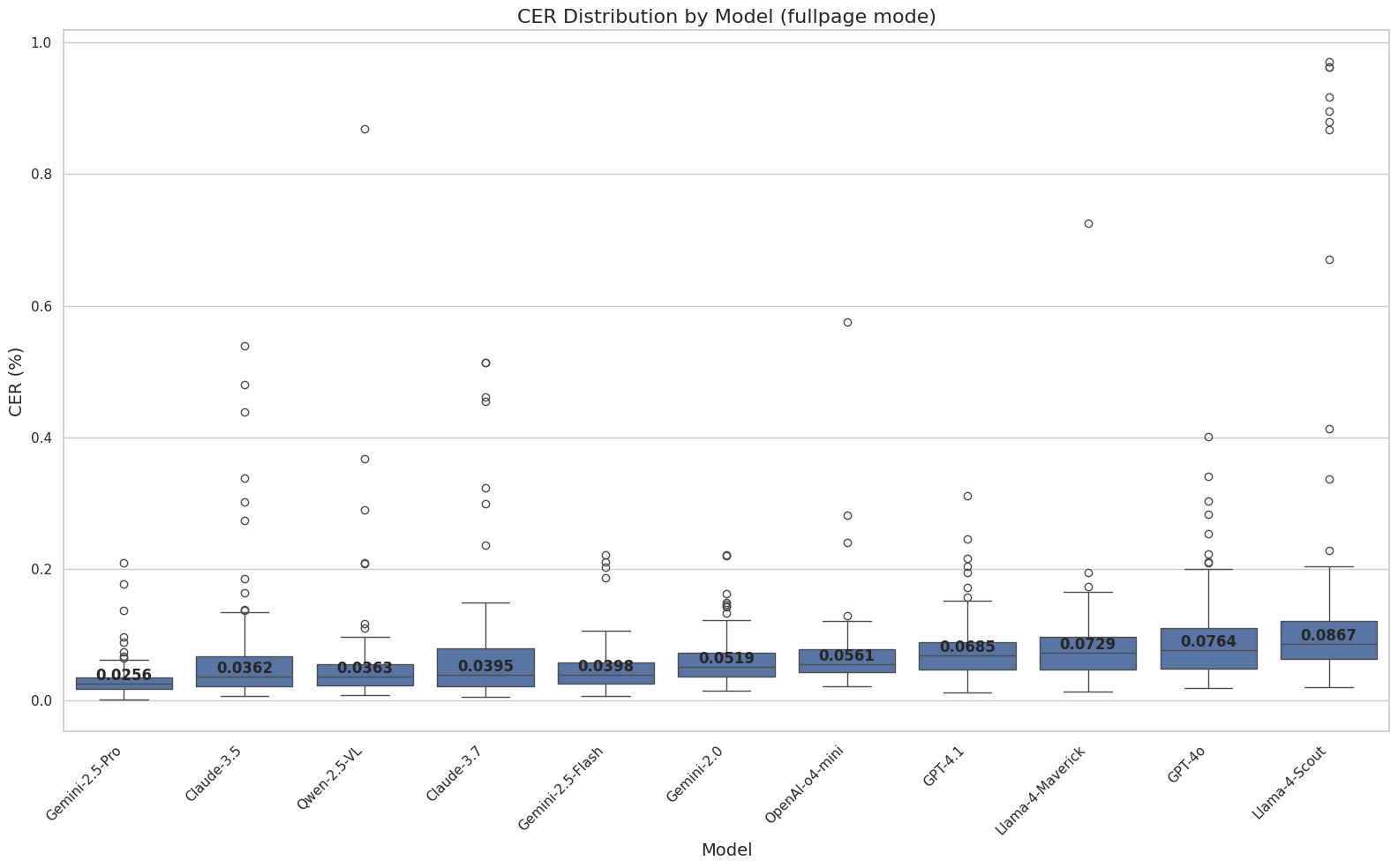}
  \caption{CER distribution by models (full page mode)}
  \label{fig:cer}
\end{figure}

\textbf{Stability Testing}. We assessed performance consistency by processing 20 documents daily with each model for seven consecutive days. Table~\ref{tab:stability} ranks models by the coefficient of variation (CV) of daily word accuracy; lower CV indicates greater stability. Gemini-2.5-Pro showed both the highest stability and the highest mean word accuracy, while Claude-3.5 exhibited the highest variability.
The distribution of CERs by model (Figure~\ref{fig:cer}) further illustrates these differences in stability, with boxplots indicating both the central tendency and the frequency of outlier cases for each model.

\begin{table*}[ht]
\centering
\begin{tabular}{rlccc}
\hline
Rank & \textbf{Model} & \textbf{CV} & \textbf{Mean Word Accuracy} & \textbf{StdDev} \\
\hline
1 & Gemini-2.5-Pro & 0.037 & 0.9620 & 0.036 \\
2 & Gemini-2.0-Flash-Lite & 0.051 & 0.9300 & 0.048 \\
3 & Gemini-2.5-Flash & 0.081 & 0.9430 & 0.077 \\
4 & GPT-4.1 & 0.113 & 0.9044 & 0.102 \\
5 & o4-mini & 0.118 & 0.9070 & 0.107 \\
6 & GPT-4o & 0.227 & 0.8620 & 0.195 \\
7 & Claude-3.7 & 0.271 & 0.8486 & 0.230 \\
8 & Claude-3.5 & 0.307 & 0.8340 & 0.256 \\
\hline
\end{tabular}
\caption{Models ranked by output stability over seven days (lower CV = higher stability).}
\label{tab:stability}
\end{table*}

No model’s daily performance deviated by more than one standard deviation from the previous day, suggesting overall day-to-day consistency.

\begin{figure}[t]
  \includegraphics[width=\columnwidth]{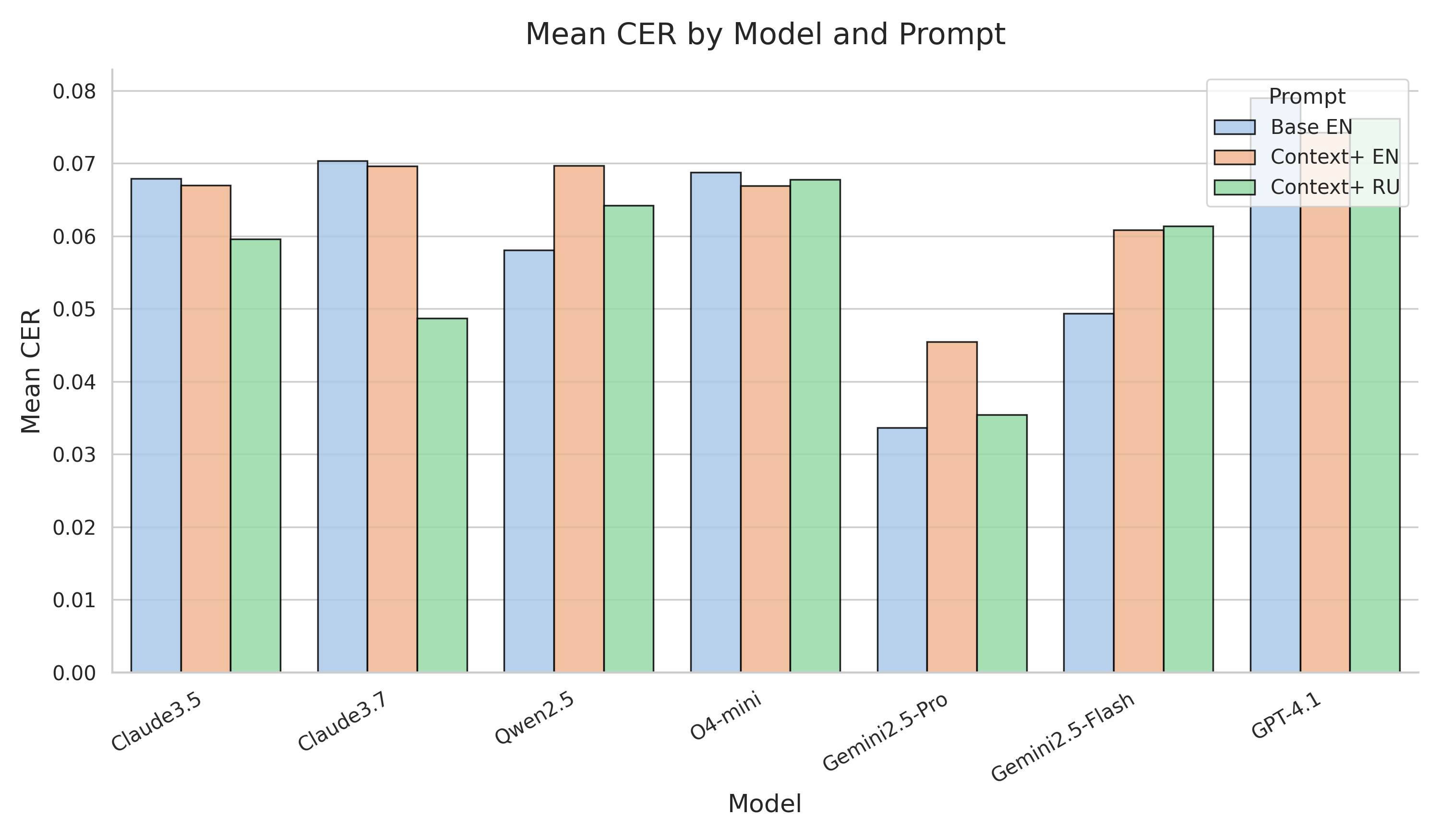}
  \caption{Mean character error rate (CER) by model and prompt strategy (simple English, context-enhanced English, context-enhanced Russian). Lower values indicate better performance.}  \label{fig:prompt}
\end{figure}

\textbf{Prompt Engineering Impact}. For the top-performing models in full-page mode, we tested three prompting strategies: a simple English prompt, a context-enhanced English prompt, and a context-enhanced Russian prompt. Context-enhanced Russian prompts led to statistically significant CER and WER reductions for several models (e.g., Claude-3.7, Claude-3.5, Gemini-2.5-Flash), with mean CER reductions of up to 0.02 and WER reductions of up to 0.03 (p < 0.05, paired t-test). Models such as GPT-4.1 and o4-mini were less affected by prompt type, suggesting greater robustness. In some cases, context-enhanced English prompts increased error rates, underscoring the impact of prompt language and structure on performance. Figure~\ref{fig:prompt} illustrates the mean CER for each model and prompt type, demonstrating the relative gains (or lack thereof) from prompt engineering across systems.

\begin{figure*}[t]
  \includegraphics[width=\textwidth]{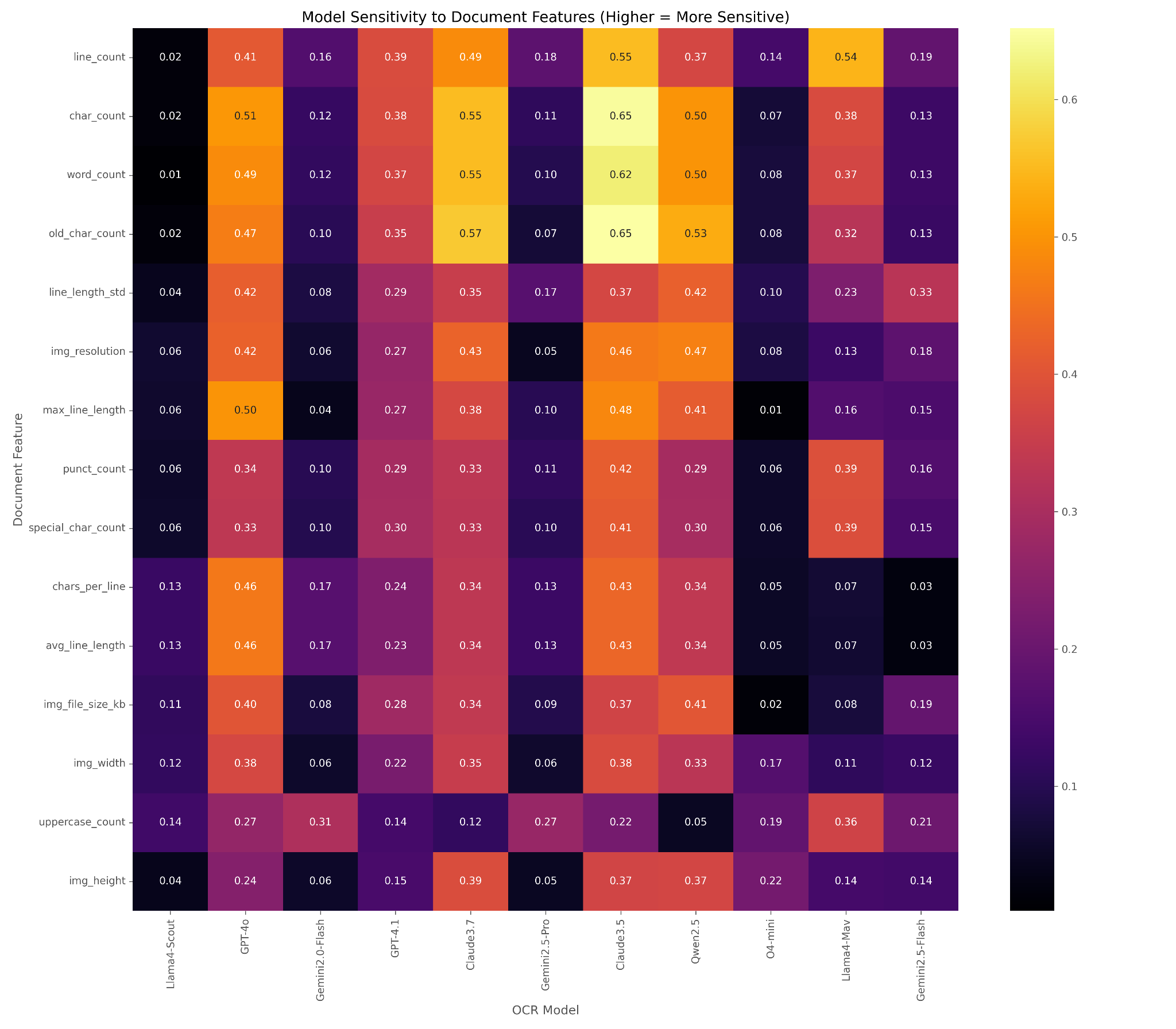}
  \caption{Model sensitivity to document features. Each cell shows the absolute correlation between a given feature (rows) and OCR error rates (CER/WER, averaged) for each model (columns; names shortened for readability). Higher values indicate greater sensitivity—that is, a model’s performance degrades more as that document feature increases. The most robust models (e.g., Gemini-2.5-Pro, o4-mini) exhibit consistently low sensitivity, while others (e.g., Claude3.5, Llama4-Mav) show heightened sensitivity to line count, old-character content, and layout complexity.}  \label{fig:sensitivity}
\end{figure*}

\textbf{Parameter Impact Visualization}. We quantified each model’s sensitivity to 17 document and image features by computing the absolute correlation of each feature with CER and WER (see Figure ~\ref{fig:sensitivity}). The most robust models (Gemini 2.5-Pro, Gemini 2.5-Flash, Qwen 2.5) achieved the lowest overall error rates and demonstrated the lowest sensitivity to layout complexity and line count—features that most strongly predict increased error for weaker models (e.g., Claude 3.5, Llama4-Mav). Notably, while most models were only moderately affected by old-character content, layout complexity ($r$ up to 0.39) and line count ($r$ up to 0.55) sharply increased error rates for several models. Regression analysis confirmed that text features explain the majority of variance in error ($R^2$ up to 0.83), with image features only adding modest predictive power.

Document features such as line count and layout complexity are the most predictive of model errors, and only the top-performing models demonstrate resilience to these challenges.

\textbf{Error analysis}. A striking and unexpected finding is that LLMs consistently "over-historicize" 18th-century Russian texts by inserting archaic Slavonic characters that had already been eliminated by Peter the Great's reforms. Instead of modernizing texts (the expected error direction), models frequently introduced obsolete characters, suggesting a systematic bias.
\begin{table*}[ht]
\centering
\begin{tabular}{lll}
\hline
\textbf{Model} & \textbf{Top Archaic Insertions} & \textbf{Most Common Errors} \\
\hline
GPT-4o              & ѡ, ѧ, ꙋ, ѿ, ꙗ   & ї~$\rightarrow$~і,\ ъ~$\rightarrow$~ь,\ т~$\rightarrow$~ш \\
GPT-4.1             & ѡ, ѧ, ꙋ, ꙗ      & ї~$\rightarrow$~і,\ ъ~$\rightarrow$~ь,\ т~$\rightarrow$~ш \\
o4-mini             & ѧ, ѭ, ѥ         & ї~$\rightarrow$~і,\ ъ~$\rightarrow$~ь,\ ѣ~$\rightarrow$~е \\
Gemini-2.5-Flash    & ѧ, ѥ            & ї~$\rightarrow$~і,\ т~$\rightarrow$~ш,\ ъ~$\rightarrow$~ь \\
Claude-3.7          & Minimal archaic insertions & ъ~$\rightarrow$~Ъ,\ ї~$\rightarrow$~і,\ ь~$\rightarrow$~ъ \\
Qwen2.5             & Minimal archaic insertions & ї~$\rightarrow$~і,\ т~$\rightarrow$~п,\ ъ~$\rightarrow$~ь \\
\hline
\end{tabular}
\caption{Archaic character insertions and most common OCR errors by model.}
\label{tab:archaic-insertions}
\end{table*}

Table~\ref{tab:archaic-insertions} summarizes both the top archaic character insertions and the most frequent error types for each model. While OpenAI and Gemini models are prone to introducing pre-Petrine archaic letters, all models struggle most with the preservation of ‘ї’ and accurate handling of the hard sign ‘ъ’.

\textbf{Over-historicization} appears most prominently in OpenAI models, with GPT-4o inserting archaic characters in 59\% of files. These insertions are not random, but follow recognizable patterns:

\textit{Medieval Slavonic characters}: 'ѧ' (little yus), 'ѡ' (omega), 'ꙋ' (monograph uk), and 'ѿ' (ot) were standard in medieval manuscripts but had been eliminated from Civil font by the mid-18th century.

\textit{Context-sensitive insertions}: Models insert archaic characters in predictable linguistic contexts—'ѧ' typically replaces 'я' in reflexive verb endings and after palatalized consonants, 'ѡ' appears in prepositions and prefixes, and 'ꙋ' substitutes for 'у' in specific word positions.

\textbf{Over-Complication with Diacritics}. Models often insert diacritical marks and combining characters that are not present in 18th-century Civil font, further complicating the transcription and introducing anachronistic features. This tendency may be exacerbated by visual noise and typographic ambiguity in the source material. For example, faded ink, paper discoloration, or ink bleed-through can produce artifacts that models misinterpret as diacritics or additional marks. Similarly, nonstandard or worn-out typefaces might blur the distinction between basic characters and diacritical elements, especially for visually similar Cyrillic forms.

\textbf{Character Preservation and Confusion}. Distinct error patterns are evident in the handling of period-specific characters:

‘ї’ vs. ‘і’: Although ‘ї’ is legitimate in 18th-century Civil font, models frequently replace it with ‘і’: the most common substitution error across all models.

‘ѣ’ \textit{(yat) preservation}: Rates vary widely, from 77.30\% (Claude-3.5) to 89.03\% (Gemini-2.0).

\textit{Hard/soft sign confusion}: All models have trouble with the terminal hard sign ‘ъ’—commonly omitted (Claude), replaced with ‘ь’ (Gemini), or incorrectly capitalized ‘ъ→Ъ’ (Claude-3.5/3.7).

\begin{figure*}[t]
  \centering
  \includegraphics[width=0.98\textwidth]{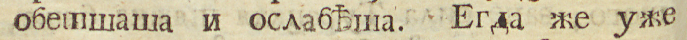}\\[2pt]
  \includegraphics[width=0.98\textwidth]{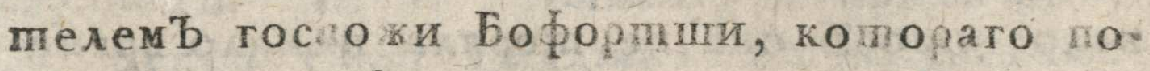}\\[2pt]
  \includegraphics[width=0.98\textwidth]{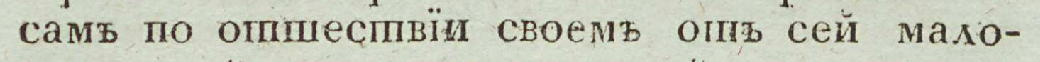}
  \caption{Excerpt from an 18th-century Russian book printed in Civil font. The letters “ш” (as in обетшаша, ослабѣша, Бофортши, отшествїи) display notable typographic variability, occasionally resembling the “т” glyph. Such variability, inherent to period printing, contributes to frequent “т→ш” substitution errors.}
  \label{fig:three_lines}
\end{figure*}

\textbf{Visual similarity errors}: Certain character pairs are frequently confused due to visual similarity—‘т→ш’ (Gemini), ‘т→п’ (Qwen, o4-mini). This confusion is exacerbated not only by scan degradation or low resolution, but also by the nature of 18th-century typography. Figure~\ref{fig:three_lines} illustrates how the Civil font renders “т” and “ш” in ways that may appear nearly identical.

These systematic error patterns offer key insights into LLM behavior on historical text:
\textbf{1) Temporal conflation}:  Models conflate orthographic features from different periods (pre-Petrine Church Slavonic, 18th-century Civil font, modern Russian), struggling to maintain strict period boundaries; \textbf{2) Contextual over-fitting}: There is a correlation between text subject and error types; models seem to apply different orthographic standards by genre, likely reflecting biases in their training data; \textbf{3) Model family signatures}:  Error profiles differ by provider (OpenAI, Anthropic, Google), suggesting differences in training data and strategies regarding historical texts.

\section{Discussion}
\textit{LLM Behavior: Over-Historicization and Error Patterns}. A surprising result is that LLMs systematically "over-historicize" 18th-century Russian texts, introducing archaic Slavonic characters that had been eliminated by the era in question. Rather than modernizing spelling, models often default to pre-reform or even medieval forms.

This likely reflects how LLMs, lacking explicit period awareness, generalize from a noisy mixture of training data: rare or visually distinctive archaic forms become signals for “historical text” regardless of actual period accuracy. Multimodal and text-only corpora contain heterogeneous historical Russian (Church Slavonic, pre-reform, post-1918), but models lack explicit period tags, so “historic” cues (yus letters, omega, diacritics) become generic signals for “old text”. Visual ambiguity in the typography and degraded print quality may further reinforce these mistakes, with models erring on the side of complexity and inserting diacritics or combining marks absent from authentic 18th-century Civil font.

\textit{Optimal Model and Prompting Strategies for Historical OCR}. Our results indicate that Gemini and Qwen models are the most robust and accurate models across diverse document types, especially where high line counts or layout complexity would otherwise increase error rates. Prompt engineering can enhance performance, particularly when prompts specify period features or are given in Russian, but the best models are less dependent on prompt tweaks. Full-page mode generally yields the best accuracy, but for models highly sensitive to document length, line-by-line mode can be preferable.

\textit{Post-OCR Correction Analysis}. Our experiments reveal counterintuitive findings about LLM post-correction effectiveness: when providing both image and OCR text to higher-performing models, performance does not exceed the correcting model's direct OCR capabilities—models essentially re-perform OCR rather than correct the provided text. We suggest that models rarely apply constrained edits; instead they re-decode from the image, using the text as weak context. Text-only correction (without source images) consistently degrades performance, with models introducing errors that corrupt the original transcription. Two mechanisms likely apply: (i) attention dilution/position effects—LLMs are known to unevenly use long contexts; (ii) editor vs. generator mismatch—chat-tuned models prioritize fluent regeneration over minimal edits unless decoding is constrained. These findings suggest practitioners should focus on selecting optimal models for direct OCR rather than post-correction pipelines, as correction attempts either provide no benefit or actively harm accuracy.

\section{Conclusion}
This paper introduces a comprehensive methodological framework for evaluating large language models (LLMs) on historical OCR tasks, exemplified by the case study of 18th-century Russian prints in Civil font. Our results demonstrate that LLM-based approaches substantially outperform traditional OCR systems for these challenging materials, and our work sets out best practices for reliable evaluation and practical implementation. Our stratified coverage across printers, decades, genres, and layouts supports transfer to other historical prints with period-specific orthography; applying the same protocol with a collection-specific grapheme inventory for HCPR/AIR typically requires only a brief 10--20-page pilot.

Looking ahead, we note that LLMs are rapidly improving; having a clearly defined evaluation protocol, public metrics, and detailed error analysis will allow ongoing, transparent tracking of model progress. However, the publication of ground-truth datasets for evaluation is a double-edged sword: once released, they risk being incorporated into future model training, compromising their utility for truly unseen evaluation. Even a single benchmarking release may affect evaluation integrity if outputs are shared or scraped. The trade-off between transparency, reproducibility, and long-term benchmark validity remains an open question for the community.

\section{Limitations}

Our dataset is specific to Russian Civil font print from the second half of the 18th century, and our manual ground truth verification process, while rigorous, may still be subject to rare annotation errors, especially for visually ambiguous or degraded source material. All evaluated LLMs were accessed via their respective APIs; however, we excluded OpenAI o3 due to prohibitive usage costs. For consistency, we requested structured (JSON) outputs when supported (e.g., OpenAI models) and programmatically extracted lines from unstructured outputs otherwise. Alternative output formats, such as Markdown or raw text, may yield different recognition results and could be further investigated in future work. Additionally, our model stability experiments revealed that LLM outputs can vary between runs for the same document and model, though this variance was relatively minor within our observation window. Nonetheless, this inherent non-determinism may affect reproducibility and should be considered when interpreting comparative results.

During evaluation, if a model’s response consisted of a clear API error message (e.g., “Unable to process image” or an explicit failure code), we resubmitted the OCR request to ensure that temporary API or service issues did not affect the results. However, if a model returned a plausible but off-target output (such as an explanation, commentary, or unrelated generative text instead of a transcription), we recorded this as the model’s result without resubmission, in line with our goal of measuring real-world output quality rather than optimizing for best-case scenarios.

\bibliographystyle{acl_natbib}
\bibliography{ranlp2025.bib}

\appendix
\section{Evaluated Models}
\label{sec:models}

\begin{table}[ht]
\centering
\begin{tabular}{|p{2.3cm}|p{4.4cm}|}
\hline
\textbf{Provider} & \textbf{Model Name} \\
\hline
Anthropic     & Claude 3.7 Sonnet \\
Anthropic     & Claude 3.5 Sonnet 20241022 \\
OpenAI        & GPT-4o-2024-08-06 \\
OpenAI        & GPT-4.1-2025-04-14 \\
OpenAI        & o4-mini-2025-04-16 \\
Google  & Gemini 2.0 Flash \\
Google        & Gemini 2.5 Pro (05-06) \\
Google        & Gemini 2.5 Flash (04-17) \\
Google  & Gemini 2.0 Flash-Lite \\
Qwen AI       & Qwen2.5-VL-72B-Instruct \\
Meta          & Llama-4 Maverick 17B 128E Instruct FP8 \\
Meta          & Llama-4 Scout 17B 16E Instruct \\
\hline
\end{tabular}
\label{tab:evaluated-models}
\end{table}

\section{Dataset Description}
\label{sec:dataset}

The evaluation dataset comprises 1,029 page images sampled from 428 unique Russian books published between 1752 and 1801, with the majority printed in the 1780s and 1790s. The collection covers a broad range of genres, with the largest shares contributed by fiction (22.7\%), religion (15.7\%), history (15.0\%), and science (12.9\%). This diversity helps ensure that both typographical and linguistic variation in Russian print is well-represented for OCR evaluation. All texts are printed in the Civil font, introduced by Peter the Great's typographic reform.

\begin{figure*}[t]
  \includegraphics[width=\textwidth]{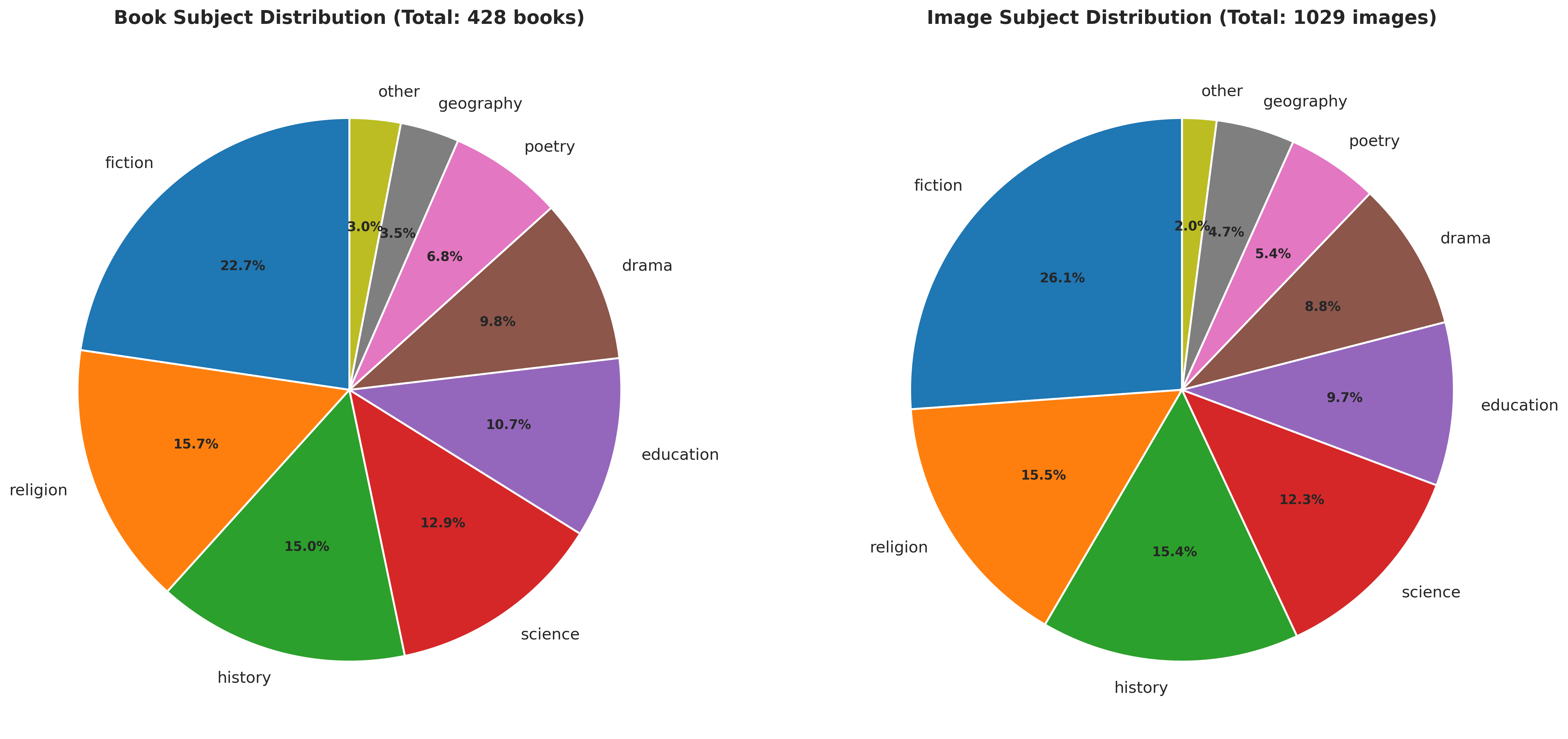}
  \caption{
  Subject distribution in the evaluation dataset.
  \textbf{Left:} Distribution by unique books (N=428). \textbf{Right:} Distribution by sampled page images (N=1029).
  The dataset is dominated by fiction, religion, history, and science, but maintains coverage across a variety of genres, supporting generalizable evaluation of historical OCR models.
  }
  \label{fig:subject}
\end{figure*}

\begin{figure}[t]
  \includegraphics[width=\columnwidth]{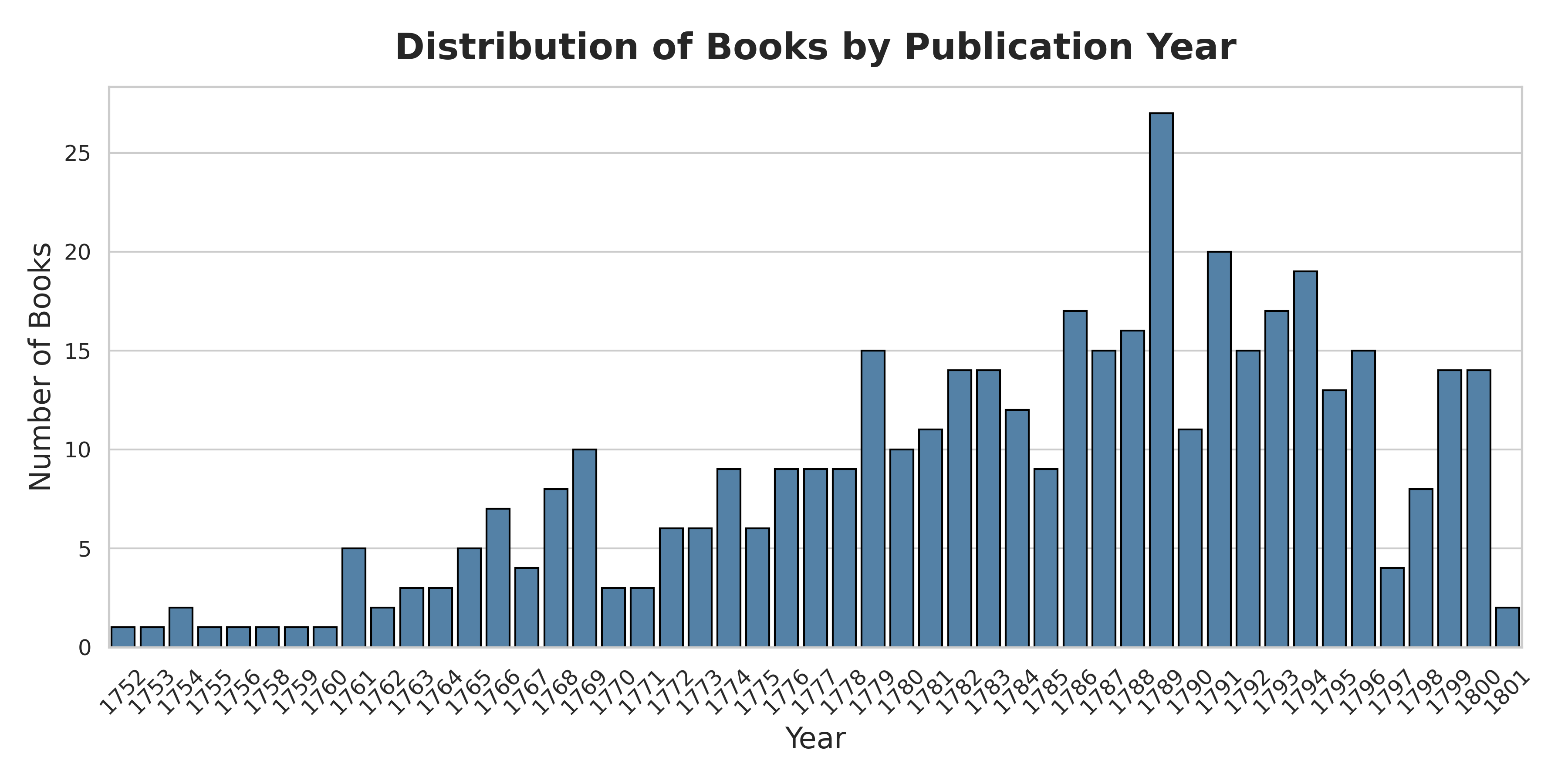}
  \caption{Year distribution in the evaluation dataset. The corpus reflects the rapid growth of Russian print in the late 18th century.}
  \label{fig:year}
\end{figure}

The resulting corpus contains 28,657 lines and 146,690 words. The 100-page sample from the corpus is published online alongside the LLM-based OCR results (\href{https://github.com/mary-lev/historical-ocr-analysis}{github repository} contains ground truth transcriptions and model outputs).
The year and subject distributions are shown in Figures~\ref{fig:subject} and \ref{fig:year}. The dataset is dominated by fiction, religion, history, and science, but maintains coverage across a variety of genres, supporting generalizable evaluation of historical OCR models

\section{Prompts}
\label{sec:prompts}

\textbf{Single Line Mode Prompt.} Extract the OCR text from this 18th-century Russian book line. Preserve the original Old Russian orthography. Respond with ONLY a JSON object containing the extracted text in the \texttt{'line'} field.

\textbf{Sliding Window Mode Prompt.} Extract the text from these consecutive lines of an 18th-century Russian book. Focus on the middle line while using surrounding lines as context. Preserve the original Old Russian orthography. Respond with ONLY a JSON object containing the extracted text of the middle line in the \texttt{'line'} field.  
Do not include any additional explanations. 

\textbf{Full Page Mode Prompt.} Extract the OCR text from this full page of an 18th-century Russian book. Preserve the original Old Russian orthography. Process each line independently. Respond with ONLY a JSON array where each object has a \texttt{'line'} field containing the transcribed text. Do not include any additional explanations.

\textbf{Full Page Context-Enhanced Prompt (English).} You are an expert OCR system specialized in processing 18th-century Russian texts. Your task is to accurately transcribe text from an image of a page from a \{book\_year\} Russian book titled ``\{book\_title\}'' published in \{publication\_info\}.  

Instructions:

Analyze the entire image thoroughly before beginning transcription.

Process the text line by line, maintaining the exact layout of the original page.

Preserve all original Old Russian orthography, including:  

-- special characters: ѣ, ѳ, ѵ, і, ї, ъ  

-- Original punctuation  

-- Capitalization as it appears in the original text.

Respond with ONLY a JSON array where each object has a \texttt{'line'} field containing the extracted text.  
Do not include any explanations or additional formatting in your response.

\textbf{Full Page Context-Enhanced Prompt (Russian).} Вы являетесь экспертной OCR-системой, специализирующейся на обработке русских текстов XVIII века, напечатанных гражданским шрифтом после реформы Петра I (1708–1710 гг.), но до реформы орфографии 1918 года. Ваша задача — точно транскрибировать текст с изображения страницы из русской книги \{book\_year\} года под названием ``\{book\_title\}'', опубликованной в \{publication\_info\}. Особенности орфографии этого периода включают:

Наличие специфических букв: ѣ (ять), і (и десятеричное) или ї, ѳ (фита), ѵ (ижица), ъ (твёрдый знак на конце слов)

Отсутствие букв церковнославянского алфавита (ѡ, ѧ, ѱ, etc.)

Использование гражданского шрифта вместо устава или полуустава

Инструкции:

Тщательно проанализируйте всё изображение перед началом транскрипции.

Обрабатывайте текст построчно, сохраняя точное расположение оригинальной страницы.

Сохраняйте всю оригинальную старорусскую орфографию, включая:  

-- специальные символы: ѣ, ѳ, ѵ, і, ї и ъ,  

-- оригинальную пунктуацию,  

-- заглавные буквы так, как они представлены в оригинальном тексте.

Отвечайте ТОЛЬКО JSON-массивом, где каждый объект имеет поле \texttt{'line'}, содержащее каждую извлеченную строку текста. Не включайте никаких пояснений или дополнительного форматирования в ваш ответ.
\end{document}